\documentclass[journal,twoside,web]{ieeecolor}
\usepackage{lcsys}
\usepackage{cite}
\usepackage{amsmath,amssymb,amsfonts}

\usepackage{graphicx}
\usepackage{textcomp}
\def\BibTeX{{\rm B\kern-.05em{\sc i\kern-.025em b}\kern-.08em
    T\kern-.1667em\lower.7ex\hbox{E}\kern-.125emX}}
\usepackage{lmodern}
\usepackage{physics}
\usepackage{xargs}                        
\usepackage{babel,csquotes,xpatch}
\usepackage[pdftex,dvipsnames]{xcolor}  
\usepackage{algorithm}
\usepackage{algpseudocode}
\newtheorem{Remark}{Remark}
\usepackage{tikz}
\usepackage{textcomp}
\usepackage{hyperref}
\usepackage{lipsum}
\newcommand\copyrighttext{%
	\tiny \textcopyright 2021 IEEE.  Personal use of this material is permitted.  Permission from IEEE must be obtained for all other uses, in any current or future media, including reprinting/republishing this material for advertising or promotional purposes, creating new collective works, for resale or redistribution to servers or lists, or reuse of any copyrighted component of this work in other works.
	DOI: \href{https://doi.org/10.1109/LCSYS.2021.3135835}{10.1109/LCSYS.2021.3135835}}
\newcommand\copyrightnotice{%
	\begin{tikzpicture}[remember picture,overlay]
		\node[anchor=south,yshift=10pt] at (current page.south) {\fbox{\parbox{\dimexpr\textwidth-\fboxsep-\fboxrule\relax}{\copyrighttext}}};
	\end{tikzpicture}%
}
\begin{document}
\title{Multiple shooting for training neural differential equations on time series}
\author{{Evren Mert Turan and Johannes Jäschke}
	\thanks{This work was supported by the Norwegian Research Council through the
		AutoPRO project.}
	\thanks{E.M. Turan and J. Jäschke are with the Department of Chemical Engineering, Norwegian University of Science and Technology (NTNU), Trondheim, Norway. evren.m.turan@ntnu.no, johannes.jaschke@ntnu.no}}%

\maketitle
\copyrightnotice
\thispagestyle{empty} 
\begin{abstract}
Neural differential equations have recently emerged as a flexible data-driven/hybrid approach to model time-series data. This work experimentally demonstrates that if the data contains oscillations, then standard fitting of a neural differential equation  may result  in a ``flattened out'' trajectory that fails to describe the data. We then introduce the multiple shooting method and present successful demonstrations of this method for the fitting of a neural differential equation to two datasets (synthetic and experimental) that the standard approach fails to fit. Constraints introduced by multiple shooting can be satisfied using a penalty or augmented Lagrangian method. 
\end{abstract}

\section{Introduction}
\IEEEPARstart{M}{echanistic} or first principle modelling of systems described by differential equations requires specification of the functional form, following which parameter estimation can be undertaken given data. Neural differential equations (DEs) are a data-driven approach to developing  dynamic models from time series data. Neural DEs give continuous dynamics, allow for irregular/incomplete time series, and can be more efficient than neural network approaches through the use modern of ODE solvers \cite{Chen2018,Rackauckas2020}. In comparison to mechanistic modelling, neural DEs reduce the need to decide on functional form, while still allowing domain knowledge to be included in the model \cite{Chen2018,Rackauckas2020}.

Despite the application of neural DEs to complex problems \cite{Rackauckas2020}, there are still challenges in their use. 
The standard approach to fitting the neural DE is to iteratively calculate a trajectory by integrating the system to the final time and updating the parameters based on this trajectory. In the dynamic optimization literature, this is known as single shooting as a single trajectory is calculated that depends entirely on the parameters and initial point \cite{Biegler2010}.

Fitting a neural DE via single shooting to a system or time series with oscillatory behaviour or with a long time span can be difficult. The optimization of neural ODE, with randomly initialised weights, may result in a ``flattened out'' or low frequency trajectory that does not describe higher frequency responses, as shown in Figure \ref{fig:single-shooting-fitted-curve}.  Indeed, researchers have demonstrated that neural networks have a spectral bias: low-frequency components of functions are learnt faster during training via gradient descent  \cite{Rahaman2019,Xu2019}.

The contribution of this work is to propose and numerically demonstrate the use of the multiple shooting method to fit neural differential equations, with constraints satisfied by an Augmented Lagrangian method. In multiple shooting the idea is to form several successive time intervals from the original time span and apply single shooting to each interval. This allows for an initially discontinuous trajectory to be formed early in the optimisation. As the optimization proceeds, the trajectory becomes continuous through the enforcement of shooting constraints. Multiple shooting is widely used for optimisation of ill-conditioned and unstable systems, see \cite{Biegler2010}. We demonstrate this method on a synthetic and experimental data set, that the neural DE otherwise fails to fit.

\section{Background}
\subsection{Neural differential equations}

Neural ODEs were introduced by \cite{Chen2018} to be a differential equation specified by a neural network, i.e.:

\begin{equation}\label{eq: NODE}
	\dv{x}{t} = NN(x, u, t, \theta)
\end{equation}
where $x$ are the states, $\theta$ are the neural network parameters, $u$ is an exogenous input, and $t$ is time. As the neural ODE is restricted by construction to be the solution of a differential equation, it is not a universal approximator \cite{Dupont2019}. Nevertheless, a wide range of systems in science and engineering are described by differential equations and neural ODEs allow one to fit a model to these systems, without specifying a function form for the differential equation.

Later authors demonstrated the use of neural networks in other types of differential equations, with the potential incorporation of a known functional form (neural differential equations) \cite{Rackauckas2020}, e.g.\ a first order differential equation of the form: 
\begin{equation}\label{eq: NDE}
	\dv{x}{t} = f(x,u,NN(x, u, t, \theta),t)
\end{equation}

This formulation allows first principle knowledge, such as conservation laws or relationships between quantities to be specified, while using the neural network to model unknown relationships \cite{Rackauckas2020}. Note that equation \ref{eq: NODE} is a special form of equation \ref{eq: NDE}. 

Regardless of the formulation, the problem of training the parameters of a neural DE is the same as estimating the parameters of a differential equation. If one had access to the states and time derivatives ($x$, $\dv{x}{t}$) then the parameter estimation would be a fitting problem, i.e.\ one would not have to integrate the system. However, typically only noisy measurements of some of the states are available. There are two main approaches to estimate the parameters of a differential equation. The first is to use a two stage method where first a flexible smooth function (typically spline bases) is fit to the data to provide estimates ($\hat{x}$, $\dv{\hat{x}}{t}$) which are then used to estimate the model parameters \cite{Varah1982}. This technique requires that all states are  measured, and that the estimate $\dv{\hat{x}}{t}$ is accurate. This later requirement becomes increasingly difficult to satisfy with increasing noise and sparsity of sampling. The alternative approach is to integrate the ODE and define the cost function $C$ using the measured and predicted states, typically the sum of squared errors (SSE) is used. This is the approach most often used for neural DEs and was taken from the optimal control literature \cite{Chen2018}.

\subsection{Single shooting with neural differential equations}

Despite the potential of neural DEs, a significant issue is the existence of local minima during the training procedure. For example, consider the example of fitting a neural ODE to the spiral differential equation \cite{Chen2018,Rackauckas2020,Onken2020}:
\begin{align}\label{spiral de}
	\dv{x}{t} &= Ax^3\\
	&= 
	\left(\begin{array}{cc}
		-0.1 &2.0\\
		-2.0 &-0.1\\
	\end{array}\right)x^3
\end{align}
The system is solved with $x_0 = [2., 0.]$, and $t \epsilon [0., 6.0]$ using a Runge–Kutta method \cite{tsitouras2011runge}.  Synthetic data points are recorded at $0.1$ intervals and normally distributed noise ($\mathcal{N}(0.0,0.2)$) is introduced.

We consider the task of fitting a neural ODE with a neural network with $x^3$ as input, i.e.\ $\dv{x}{t} = f_{NN}(x^3, \theta)$, using the sum of squared error as the cost function. This means that the neural network has the task of approximating $A$ in equation \ref{spiral de}. The neural network has one hidden layer of 16 nodes, $\tanh$ is used as the activation function in the input and hidden layer, and initial weights are set via Glorot initialization \cite{Glorot2010}.

\begin{figure}
	\centering
	\includegraphics[width=1.0\linewidth]{"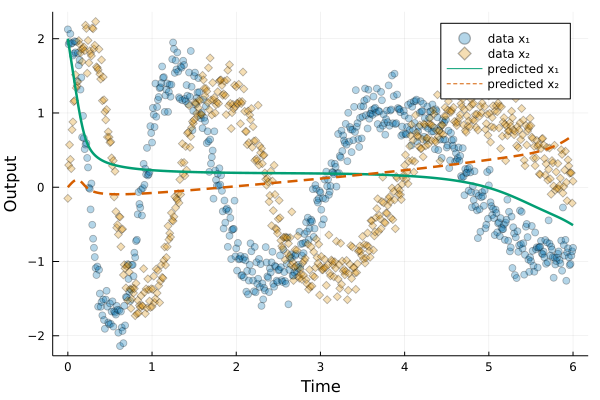"}
	\caption{Plot of neural ODE fitted to data from equation \ref{spiral de} via single shooting with Nadam \cite{Dozat2016}.}
	\label{fig:single-shooting-fitted-curve}
\end{figure}

An issue with single shooting is that the optimiser must simultaneously select parameters to improve the fit at all points along the trajectory. This can result in the network getting stuck during training by fitting a flattened curve through the middle of the data as shown in Figure \ref{fig:single-shooting-fitted-curve}. Fitting is performed by a Nesterov momentum version of the Adam algorithm (Nadam)  \cite{Kingma2014,Dozat2016}, with an initial learning rate of $0.001$. 

When changing the learning rate, different behaviour can be observed during the training. For example,  the initial oscillation of the system ($0\leq t \leq 1.5$) can be fit followed by straight or warped lines ``within'' the oscillations of the data, similar to Figure \ref{fig:single-shooting-fitted-curve}. It should be noted that the curve can be fit by careful adjustments of the learning rate (scheduling) during the optimization. However, in general this will require manual interaction.

\subsection{Multiple shooting}

\begin{figure}
	\centering
	\includegraphics[width=1.0\linewidth]{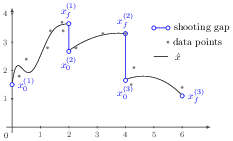}
	\caption{Schematic of multiple shooting.}
	\label{fig:multishoot}
\end{figure}

Multiple shooting is an alternative method of fitting, wherein the time span $[t_0,t_f]$ is partitioned into $N_s$ intervals by forming a grid of $N_{s}+1$ points, $t_0 = \tau_0<\tau_1<...<\tau_{N_{s}}=t_f$ \cite{Bock1981}. The values of the state $x$ at the grid points are introduced as additional variables (shooting variables) e.g.\ in Figure \ref{fig:multishoot}, interval 2 is formed by $\tau_1=2.0$ to $\tau_2=4.0$ and the initial and final values in this interval are denoted as $x_0^{(2)}$ and $x_f^{(2)}$. 

On each interval an initial value problem can be solved, giving a potentially discontinuous trajectory $\hat{x}$ as shown in Figure \ref{fig:multishoot}. This trajectory is used to calculate the cost (and gradient) as in single shooting. The trajectory becomes meaningful, when the gap between intervals (the shooting gap, see Figure \ref{fig:multishoot}) introduced by the new state variables is zero, i.e.\ at the end of the training procedure, the following constraints need to be satisfied:
\begin{equation}\label{multiple_shootin_constraint}
	x_f^{(i)} - x_0^{(i+1)} = 0 \qquad i = 1,2,\dots,N_s 
\end{equation} 

The use of multiple shooting offers two advantages for training neural DEs: (1) the time series data can be used to provide an initial guess for the unknown states at the shooting points - thus, the influence of poor initial parametrisation is reduced \cite{Bock1981}, (2) the $N_s$ initial value problems are independent and hence their solving is parallelisable. Point (1) can aid in the initial fitting of a neural DE -  while   the network weights are small (i.e. the neural network is close to linear) the optimiser can improve the fit by adjusting the shooting variables, thereby giving a discontinuous trajectory that describes the data. The disadvantage of multiple shooting is that the optimisation problem has $N_s$ constraints that need to be satisfied, e.g. by the methods outlined in the following section.

\subsection{Penalty and Augmented Lagrangian methods}
Consider the constrained optimization problem:
\begin{align}\label{eq: constrained_problem}
	&\min C(z) \\
	&\text{s.t. } h(z)=0 
\end{align}	
where $C$ is the cost function and $h$ is a vector function of equality constraints, in our case the shooting constraints (Equation \ref{multiple_shootin_constraint}), and  $z$ are the optimisation variables. 
Supervised learning of neural networks is typically performed by optimising an unconstrained problem, that is related to the constrained optimisation problem. A common approach is to define a proxy cost function, $\phi$, which has the constraints as penalties terms:  

\begin{equation}\label{eq: penalty}
	\phi=C(z) + \rho Q(h(z))
\end{equation}	
where $\rho$ is a hyper-parameter, and $Q$ is a penalty function.  The most common choices of $Q$ are a quadratic penalty function ($Q(h(z))=h(z))^2)$, or the $l_1$, $l_2$ (not squared) and $l_\infty$ norms

The later three norms give an exact penalty function, which means that under standard assumptions \cite{Nocedal2006} a single minimization  with some $\rho^*$   can yield the same solution as the constrained problem. Note that a too large $\rho$ can result in numerical issues, while a too small $\rho$ may result in constraint violation \cite{Nocedal2006}. 

An alternative approach is to use the method of multipliers or the augmented Lagrangian method which defines the objective function as:

\begin{equation}\label{eq: aug lagrangian penalty}
	\phi=C(z) +\sum h_i(z)^Tv_i+ \rho \sum h(z)_i^Th(z)_i
\end{equation}	
where $v$ is an approximation of the Lagrange multipliers, that is updated, along with $\rho$, as part of the optimization algorithm. Algorithm 1 outlines a potential Augmented Lagrangian algorithm.

\begin{algorithm}[H]
	\caption{Augmented Lagrangian}
	\begin{algorithmic}[1]
		\State Initialize the constrained optimisation problem 
		\State Set: $v\gets 0$, $\rho\gets 0$, $k\gets 1$
		\Repeat 
		\State $\theta^k,x^k= \arg\min \phi$ \Comment{Unconstrained}
		\If{$h(x^k,\theta^k)=0$}
		\State Converged $\gets$ True
		\Else
		\State Update $v$ and $\rho$ \Comment{Algorithm dependent}
		\State $k\gets k +1$ 
		\EndIf
		\Until{{Converged}}
	\end{algorithmic}
\end{algorithm}
Augmented Lagrange algorithms can use any unconstrained optimiser to solve the unconstrained optimization problem (line 4). Globally convergent augmented Lagrange algorithms have been implemented \cite{Conn1991,Birgin2008}.

\begin{Remark}
	At the time of writing, there is unpublished related work, similar in spirit, in an example in a neural differential equation package \cite{rackauckas2019diffeqflux}. In the example a penalty approach is used, and the shooting intervals are restricted to start and end on selected data points. Thus, the approach is unsuitable for real, noisy systems because the data points are noise contaminated and there is no reason why the fitted solution should go through these data points.
\end{Remark}

\section{Multiple shooting with Neural DEs}

In the following sections we demonstrate the approach on two problems. This work is coded in Julia \cite{Julia-2017}, using the following packages: DifferentialEquations.jl \cite{rackauckas2017differentialequations,rackauckas2018comparison}, DiffEqFlux \cite{rackauckas2019diffeqflux,Rackauckas2020}, Flux.jl \cite{Flux.jl-2018}, ForwardDiff.jl \cite{Revels2016}, NLopt.jl \cite{Johnson2014} and Hyperopt.jl \cite{BaggeCarlson2018}. 

\subsection{Spiral differential equation - synthetic example}

The spiral differential equation introduced in the preceding section (Eq. \ref{spiral de}) is used as a synthetic example to demonstrate the proposed procedure. The sum of squared errors (SSE) is again used for the cost function, however to allow for good out-of-sample prediction we introduce a regularisation term, $R(\theta)$, that penalises the complexity of the neural DE, i.e. $C(\theta) = \text{SSE}(\theta) + \rho_RR(\theta)$, where $\rho_R$ is a regularisation constant. 

We use the sum of the spectral norm of the weights in each layer as the regularisation term \cite{Gouk2021}, with a regularisation constant of $1.0$. Furthermore, we remove the bias nodes from the network as we wish to map zeros-to-zeros, as an application of prior knowledge. The additional variables introduced by multiple shooting are initially set to $x(0)$.

\begin{figure}
	\centering
	\includegraphics[width=1.0\linewidth]{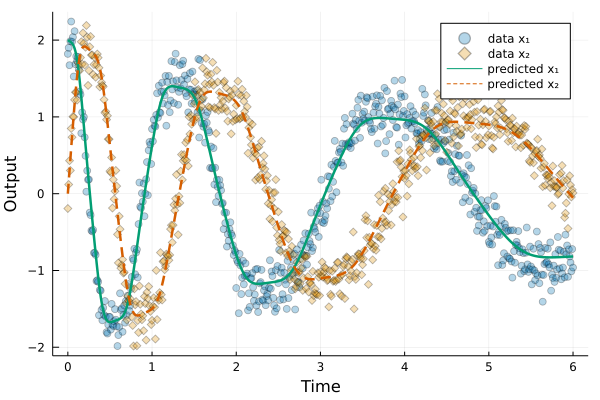}
	\caption{Plot of neural ODE fitted to data from equation \ref{spiral de} via multiple shooting with 20 intervals. Figure was made using the fitted parameters in a single IVP, i.e.\ without intervals.}	
	\label{fig:multipleshootinglipl10}
\end{figure}

\begin{figure}
	\centering
	\includegraphics[width=1.0\linewidth]{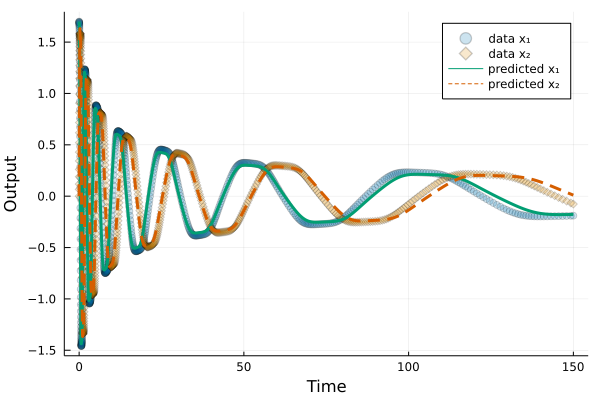}
	\caption{Plot of neural ODE fitted to data from equation \ref{spiral de} on $t \epsilon [0., 6.0]$, on a much larger span. Neural ODE fitted via multiple shooting with 20 intervals.}
	\label{fig:multipleshootingnobiasliplinf1}
\end{figure}

Training is performed with 20 intervals using an Augmented Lagrangian method \cite{Conn1991,Birgin2008}, with LBFGS \cite{Liu1989} used as the inner optimiser. Figure \ref{fig:multipleshootinglipl10} shows that in comparison to single shooting (Figure \ref{fig:single-shooting-fitted-curve}), multiple shooting is able to give a neural ODE that fits the data. Moreover, the trained neural ODE shows good generalisation to a much longer time scale  $t \epsilon [0., 250.0]$, as shown in Figure \ref{fig:multipleshootingnobiasliplinf1}, despite only being trained on data up to $t = 6.0$. This is partially because the removal of the bias nodes forces the neural DE to have a time derivative of $0.0$ when both states are $0.0$.

\subsection{Cascading tanks - real data}
\subsubsection{System description}
\begin{figure}
	\centering
	\includegraphics[width=0.75\linewidth]{"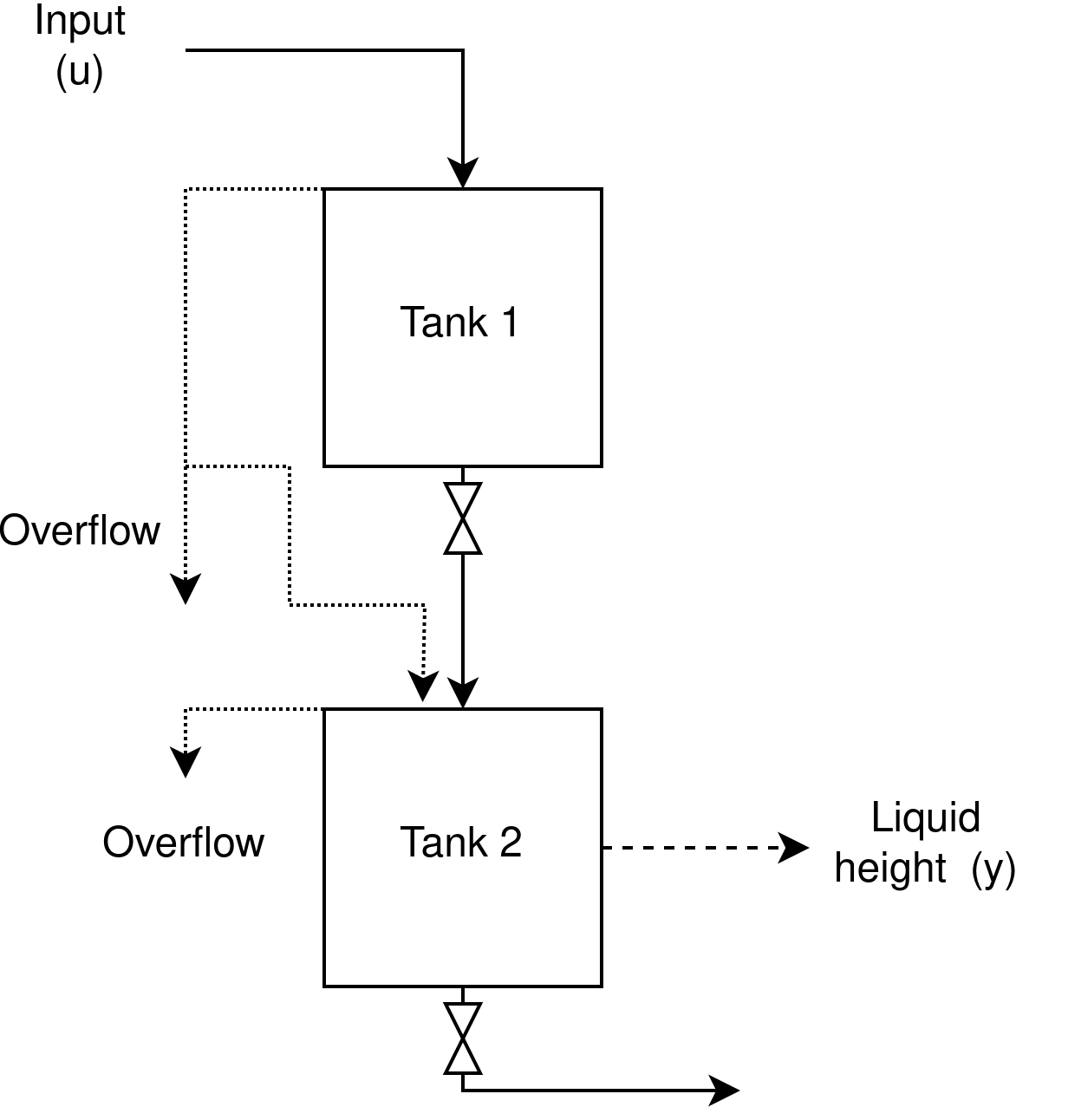"}
	\caption{Diagram of cascading tank system}
	\label{fig:diagram}
\end{figure}

The cascading tank system considered here is part of a non-linear identification benchmark problem, fully described in \cite{Schoukens2017}. The system consists of two tanks, arranged as per Figure \ref{fig:diagram}. Water is fed to tank one (the pump voltage is the exogenous input signal, $u$), and then flows into tank two before leaving the system. Water can also overflow over the edge of the tanks, and a portion of the overflow from tank one may enter tank two. Only the water level of the second tank (output signal, $y$) is recorded. 

\begin{figure}
	\centering
	\includegraphics[width=1.0\linewidth]{"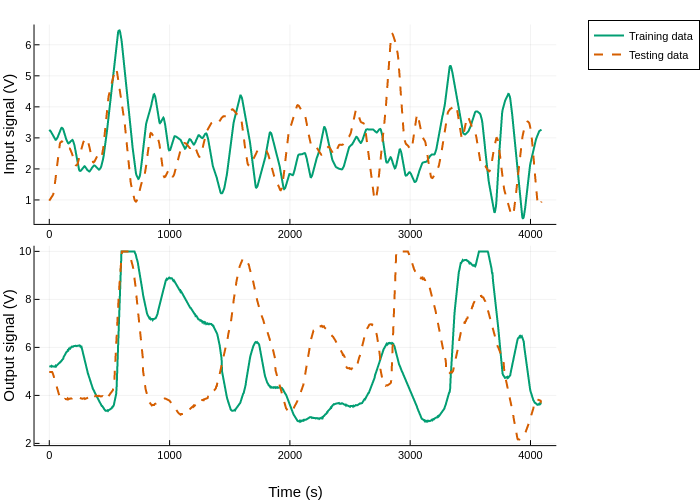"}
	\caption{Input and output signals for the cascading tanks system \cite{Schoukens2017}}
	\label{fig:data}
\end{figure}

The benchmark problem consists of a training and testing dataset. The input, $u(t)$, is a multisine signal consisting of 1024 measurements (at 4 seconds intervals), and the output signal, $y(t)$, is the measurement from the water level sensor at these same time points, giving the data shown in Figure \ref{fig:data}. The second half of training set is used for validation. The initial height of the tanks are unknown, but are the same for both data sets, i.e. $y(0)$ is estimated from the data. Our goal is to fit a neural ODE to this dataset. The SSE is used as the cost function. As the input signal ($u$) is discrete,  we use a constant piecewise interpolation of the data for evaluation in continuous time. In comparison, using a cubic spline has influence on the results. Inputs before the data period are assumed to be constant, i.e.\ $u(t)=u(0),\enspace \forall t<0.0$. 

\subsubsection{Neural network}
A neural network with no bias units, and one hidden layer with 64 units, $\tanh$ as the activation function in the input and hidden layer, and initial weights set via Glorot initialization \cite{Glorot2010}. For regularisation, an $l_2$ penalty is applied on the network weights, with the penalty constant, $\rho_{l_2}$, as a hyper parameter. 

Using only the input, $u(t)$, and output, $y(t)$, signals as features for the neural network gives a poor fit to the data. As such we provide three additional inputs:
\begin{itemize}
	\item $\sqrt{y(t)}$ - the flow out of a tank is proportional to the square root of the fluid height (Bernoulli's equation)
	\item $u(t-\tau_d)$ - the first tank acts as a time delay 
	\item $\int_{t-\tau_i}^{t}u(t^*)dt^*$  - the output signal shows less rapid variation than the input as the first tank ``smooths'' the data (see Figure \ref{fig:data})
\end{itemize}
Thus, we are fitting the neural DE:
\begin{align}
	\dv{y}{t} =& NN\bigg(u(t),y(t),\sqrt{y(t)},u(t-\tau_d),\\ &\int_{t-\tau_i}^{t}u(t^*)dt^*\bigg)\nonumber\qquad\text{$NN: \mathbb{R}^5\to\mathbb{R}$}
\end{align}

\subsubsection{Fitting}

\begin{figure}
	\centering
	\includegraphics[width=.90\linewidth]{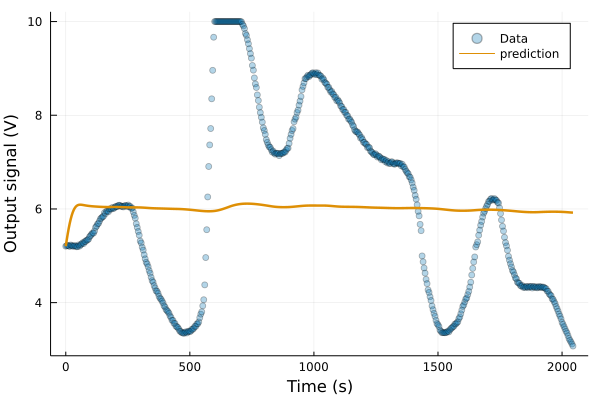}
	\caption{Neural ODE for cascading tanks system fitted via single shooting, with Adam \cite{Kingma2014} (2000 iterations, with learning rate set to 0.001).}
	\label{fig:singleshootingtestdata}
\end{figure}

\begin{figure}
	\centering
	\includegraphics[width=0.9\linewidth]{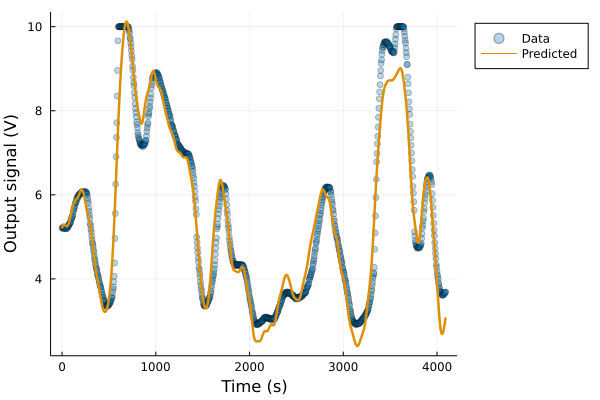}
	\caption{Neural ODE for cascading tanks system fitted via multiple shooting, with an Augmented Lagrange method \cite{Birgin2008}. The first half of the data (0-2048 seconds) was used for training, and the later half for validation (2048-4096 seconds). }
	\label{fig:cascading_train}
\end{figure}

\begin{figure}
	\centering
	\includegraphics[width=.90\linewidth]{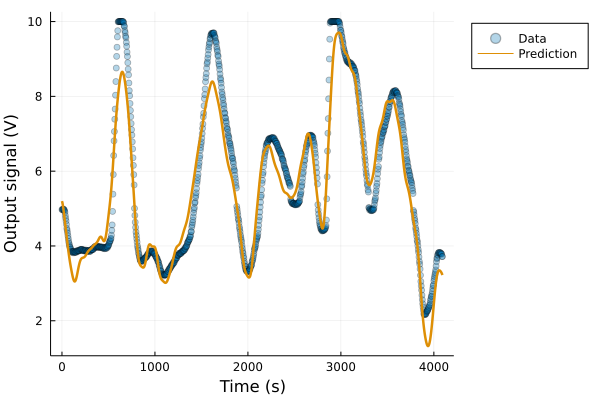}
	\caption{Neural ODE for cascading tanks system fitted via multiple shooting, with an Augmented Lagrange method \cite{Birgin2008}.}
	\label{fig:cascading_test}
\end{figure}

Fitting the neural DE via single shooting proceeds poorly as shown in Figure \ref{fig:singleshootingtestdata}. In comparison, with multiple shooting the neural ODE is able to be fit to the data (Figure \ref{fig:cascading_train}). The average square root error on the training, validation and test set is $0.42$, $0.50$, $0.62$ respectively. Figure \ref{fig:cascading_test} shows that the neural DE is able to generalise well, although it has issues with the large peaks in the first 2000 seconds. The hyper parameters values are chosen via Bayesian optimization  as $5.96\times10^{-2}$, $79.0$s, and $164.0$s, for $\rho_{l_2}$, $\tau_d$, and $\tau_{i}$  respectively. See \cite{Frazier2018} for an introduction to Bayesian optimization.

\section{Discussion}

The examples shown in the paper use the augmented Lagrangian method, however we found that using a penalty method  is feasible. The penalty parameter strongly influences the fit of the neural ODE, with potential approaches to adjust this parameter discussed in \cite{Nocedal2006} and \cite{Biegler2010}.

Authors have recently showed that the interaction between neural DE and DE solver can lead to discrete dynamics, resulting in the neural DE depending on the numerical methods used in the fitting \cite{Ott2020}.  A specialised time stepping algorithm was recommended \cite{Ott2020}, however the use of ``normal'' adaptive time stepping, as implemented in DifferentialEquations.jl \cite{rackauckas2017differentialequations} showed no issues.

\section{Conclusion}

Fitting a neural DE to a time series with oscillatory behaviour can be a challenging task. Multiple shooting can alleviate this difficulty, by providing the optimiser the flexibility to find an initially discontinuous trajectory that is close to the observed data \cite{Bock1981}. This was demonstrated through fitting a synthetic and experimental  dataset. 
An augmented Lagrangian method was used to fit the neural ODE due to the shooting interval constraints. In practice the penalty method can work well, if the penalty parameter is carefully chosen.
Future work could investigate the effect on computational time due to the introduction of constraints and the possible parallelization, or the influence of the length of the shooting intervals. 
   \bibliographystyle{ieeetr}
\bibliography{bib}

\end{document}